\documentclass[letterpaper]{article} 
\usepackage[draft]{aaai2026}  
\usepackage{times}  
\usepackage{helvet}  
\usepackage{courier}  
\usepackage[hyphens]{url}  
\usepackage{graphicx} 
\usepackage{natbib}  
\usepackage{caption} 
\usepackage{algorithm}
\usepackage{algorithmic}

\usepackage{newfloat}
\usepackage{booktabs}
\usepackage{qtree}
\usepackage[table,xcdraw]{xcolor}
\usepackage{listings}
\usepackage{makecell}
\usepackage{multirow}
\usepackage{colortbl}
\usepackage{amsmath}
\usepackage{subfigure}
\usepackage{subcaption}
 
\DeclareCaptionStyle{ruled}{labelfont=normalfont,labelsep=colon,strut=off} 
\floatstyle{ruled}
\newfloat{listing}{tb}{lst}{}
\floatname{listing}{Listing}
\title{LLMs Know More Than Words: A Genre Study with Syntax, Metaphor \& Phonetics}
\author{
    Weiye Shi\textsuperscript{\rm 1},
    Zhaowei Zhang\textsuperscript{\rm 1},
    Shaoheng Yan\textsuperscript{\rm 1},
    Yaodong Yang\textsuperscript{\rm 1}
}

\affiliations{
    \textsuperscript{\rm 1}Institute for Artificial Intelligence, Peking University\\
    Beijing, China\\
    shiweiye@stu.pku.edu.cn
}

\usepackage{bibentry}

\begin{document}

\maketitle

\begin{abstract}
Large language models (LLMs) demonstrate remarkable potential across diverse language‑related tasks, yet whether they capture deeper linguistic properties—such as syntactic structure, phonetic cues, and metrical patterns—from raw text remains unclear. To analysis whether LLMs can learn these features effectively and apply them to important nature language related tasks, we introduce a novel multilingual genre classification dataset derived from Project Gutenberg, a large-scale digital library offering free access to thousands of public domain literary works, comprising thousands of sentences per binary task (poetry vs.\ novel; drama vs.\ poetry; drama vs.\ novel) in six languages (English, French, German, Italian, Spanish, and Portuguese). We augment each with three explicit linguistic feature sets (syntactic tree structures, metaphor counts, and phonetic metrics) to evaluate their impact on classification performance. Experiments demonstrate
that although LLM classifiers can learn latent linguistic structures either from raw text or from explicitly provided features, different features contribute unevenly across tasks, which underscores the importance of incorporating more complex linguistic signals during model training.
\end{abstract}


\begin{figure*}[t]
\centering
\includegraphics[width=1.5\columnwidth]{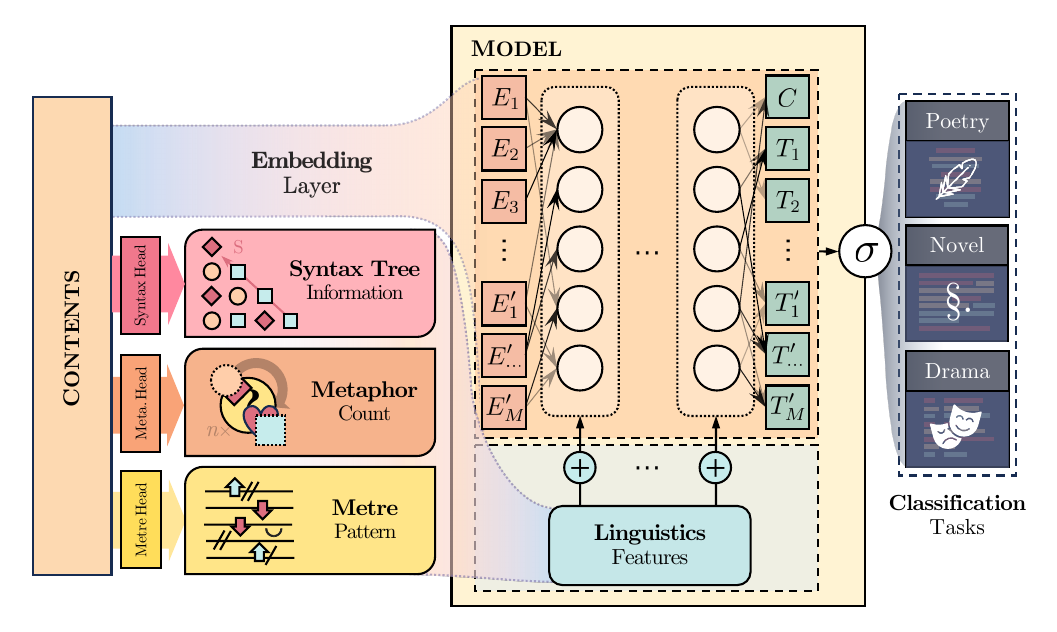} 
\caption{Overview of our method. We extract various types of linguistic information from raw text and integrate them with the original sentences. The model is then trained to embed these enriched inputs uniformly into the latent space, thereby enhancing performance on genre classification tasks.}
\label{fig1}
\end{figure*}

\section{Introduction}

Large Language Models are extensively investigated in text generation, translation, summarization, and many kinds of language-related tasks \citep{10822885, li2024personal}, along with widespread adoption across the social sciences and liberal arts \citep{thapa2025large, ziems2024can, wang2024large}. Meanwhile, researchers have begun exploring their intersections with linguistic science \citep{munoz2024contrasting,rosenfeld2024whose}. Some studies have examined whether synthetic texts generated by LLMs differ fundamentally from human‑authored texts \citep{munoz2024contrasting}, while others have sought to uncover novel linguistic patterns emerging in AI outputs \citep{kuwanto2024linguistics}, which can be used both for developing robust methods to detect LLM‑generated content \citep{park2025katfishnetdetectingllmgeneratedkorean} and for advancing automated language generation. However, despite the crucial role that linguistic features play in natural language, relatively little research has been conducted in this area. The ability of models to comprehend and leverage these features for downstream tasks has long been overlooked, leaving a significant gap in our understanding of how linguistic structures can enhance model performance. Therefore, a dataset that combines sufficient scale, linguistic complexity, and multilingual coverage is essential for training and evaluating models on these fundamental tasks. Moreover, a rigorous and scientific framework is also needed to assess how well models understand latent linguistic features.

\label{Introduction}
We argue that genre classification offers the most effective framework for evaluating a model’s grasp of linguistic features. Literary scholars have long recognized that genres are defined by distinctive linguistic patterns: they vary in syntactic tree structures \citep{dell2018lexicon, brigadoi2021genre}, in the use of metaphor, and in phonetic characteristics such as metrical patterns. By focusing on these core differences, genre classification directly tests a model’s ability to capture the latent structures that distinguish one form of writing from another.

Therefore, in this paper, we construct a multilingual genre dataset comprising novels, poetry, and drama from Project Gutenberg, a large-scale digital library offering free access to thousands of public domain literary works and always used in linguistic research, to investigate how latent linguistic structures influence a model’s ability to distinguish between literary forms. For each language (English, French, German, Italian, Spanish, and Portuguese) we extract thousands of sentences for three binary classification tasks (poetry vs.\ novel, drama vs.\ poetry, and drama vs.\ novel), splitting the data into 80\% training and 20\% test sets. We then fine-tune models using a naive approach as well as with three explicit linguistic features: syntactic tree depth, metaphor frequency, and phonetic regularity. To evaluate performance, we compute the F1 score for each binary task and compare the baseline with the linguistically informed models. Results show consistent improvements across most metrics when incorporating linguistic features. Experiments demonstrate that incorporating phonetic features into the training process yields consistent performance improvements across most models. Additionally, adding syntactic tree structures and metaphor counts produces targeted gains that align closely with the statistical properties of each genre. These results indicate that the classifiers effectively learn latent linguistic features in the sentences and that different feature types contribute variably depending on the specific classification task.

In conclusion, this paper makes three key contributions. \textbf{First}, we introduce a novel multilingual literary genre classification dataset derived from Project Gutenberg, a large-scale digital library offering free access to thousands of public domain literary works, spanning six languages and three binary genre distinctions, which serves as a fundamental tool for evaluating LLMs' sensitivity to latent linguistic features. \textbf{Second}, we systematically augment this dataset with three types of linguistically motivated features, providing a framework to assess how explicit linguistic cues affect model performance. \textbf{Third}, through extensive experiments, we demonstrate that LLMs can both infer and benefit from these latent structures, showing varied but consistent performance gains, which underscores the importance of linguistically enriched signals in advancing LLM-based literary and language understanding.

Based on the experimental results, we derive the following insights:

(1) \textbf{Incorporating latent linguistic structure information into model training can enhance performance on various language tasks that involve complex linguistic features.} This approach offers a promising pathway to align model-generated content more closely with human language, potentially bridging the subtle gaps that often arise between AI-generated and human-authored texts.

(2) \textbf{Different linguistic features contribute unevenly to model learning.} Grammatical cues such as syntactic tree structures are relatively easy for models to learn, whereas phonetic patterns are more subtle and difficult to capture, as they are closely tied to sound and thus inherently multimodal. Interestingly, these more latent features tend to yield greater performance improvements, suggesting that supplying models with deeper, multimodal linguistic signals may significantly boost their capacity to handle complex language tasks.

\section{Related Work}
In this section, we review related work on the interaction between linguistic features and large language models (LLMs), and discuss how the task of genre classification fits into this broader research landscape.

\paragraph{Linguistic Features and LLMs.} Interdisciplinary research at the intersection of large language models (LLMs) and linguistics has only begun to emerge in recent years. Early studies focused on evaluating whether encoder-only models, such as BERT \citep{koroteev2021bert}, possess the capacity to understand the construction grammar of sentences \citep{tayyar-madabushi-etal-2020-cxgbert}. Subsequent work has involved the use of LLMs to annotate sentence and paragraph structures \citep{yu2024assessing, ljubešić2024classlawebcomparablewebcorpora}, as well as comparing the linguistic characteristics of sentences generated by decoder-only models, such as LLaMA \citep{touvron2023llama}, Falcon \citep{harabagiu2000falcon} and Mistral \citep{jung2010mistral}—with those written by humans \citep{munoz2024contrasting}. Researchers have also begun to investigate whether LLMs are capable of understanding phonetic \citep{ballier2023using,doshiphonetic} and morphological information \citep{anderson2025unsupervised,asgari2025morphbpe} embedded in sentences. However, no study has exclusively focused on how linguistic characteristics can be measured through genre classification tasks. Moreover, existing datasets and benchmarks have primarily concentrated on English texts and often lack rigorous validation based on linguistic theory. These gaps highlight the need for a multilingual and theoretically grounded evaluation pipeline to assess how effectively models classify genres and to provide deeper insights into the role of underlying linguistic structures in this fundamental task.

\paragraph{Genre Classification.} Genre classification, by contrast, has long been explored through computational methods. Early efforts focused on using statistical learning techniques to automatically classify genres based on the lexical features of documents \citep{10.1145/564376.564403}. In the era of large language models (LLMs), most research has centered on constructing datasets from a functional perspective \citep{https://doi.org/10.1002/asi.23308, kuzman2022gincotrainingdatasetweb, doi:10.3366/cor.2018.0136, make5030059}, distinguishing between genres such as news articles, speeches, fiction, and song lyrics. Other work has examined emotional or stylistic dimensions \citep{hicke2025lookinginnermusicprobing}. However, the prevailing methodologies still rely largely on word-level features \citep{bhattacharjee2024llmguidedcausalexplainabilityblackbox}, and tend to focus on monolingual or limited multilingual contexts, often overlooking the broader multilingual dimension \citep{kuzman2022gincotrainingdatasetweb}.

\section{Method}
In this section, we introduce the construction of our dataset, outline our problem formulation, and explain how linguistic features are incorporated into the training process.

\subsection{Data Creation}
\label{sec:dataset}
Our dataset is derived from Project Gutenberg, a digital repository of public-domain e-books that are free from copyright restrictions. We sampled approximately 1500-3000 sentences for each of six languages, English (EN), French (FR), German (DE), Spanish (ES), Italian (IT), and Portuguese (PT), across three canonical literary genres: drama, poetry, and the novel. The exact number depends on the scale of the available raw text.This tripartite genre classification is well-established in literary scholarship. In total, the dataset comprises roughly 45,000 sentences. For each language–genre subset, we split the data into training and testing sets using an 80/20 ratio.

Next, we constructed three binary classification tasks—poetry vs. novel, poetry vs. drama, and novel vs. drama—by pairing each two genres. This setup enables us to more clearly investigate how different linguistic features influence genre classification across varying levels of structural and stylistic contrast. A detailed breakdown of the dataset distribution is provided in Table~\ref{tab:dataset}.

\begin{table}[ht!]
\centering
\caption{Dataset statistics by genre and language.}
\label{tab:dataset}
\vspace{-0.5em}
\resizebox{0.3\textwidth}{!}{\begin{tabular}{lccc}
\toprule
Language & Drama & Poetry & Novel \\
\midrule
EN & 1625 & 3367 & 2633 \\
FR & 2313 & 2092 & 2397 \\
DE & 2528 & 2443 & 3481 \\
ES & 2423 & 2795 & 3102 \\
IT & 1912 & 2474 & 2836 \\
PT & 1658 & 1530 & 2734 \\
\bottomrule
\end{tabular}}
\vspace{-0.5em}
\end{table}

\begin{figure*}[t]
\centering
\includegraphics[width=0.8\textwidth]{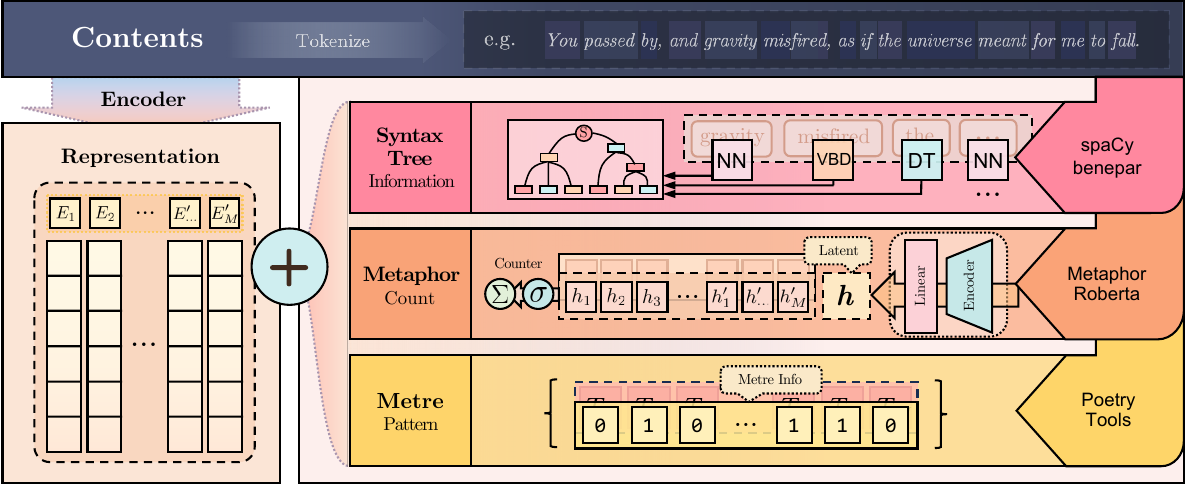} 
\caption{Detailed process of our method. We extract syntactic tree information and metrical patterns using two well-established natural language processing tools: spaCy-Benepar and PoetryTools. Metaphor counts are obtained using Metaphor RoBERTa, a state-of-the-art pretrained model designed to detect word-level metaphor usage. During training, we integrate the original sentence with the extracted linguistic features to enhance the model’s performance.}
\label{fig2}
\end{figure*}

\subsection{Embedding Linguistic Features}
As mentioned in Introduction\ref{Introduction}, we hypothesize that several linguistic features may assist LLMs in solving the genre classification task. Lyrical genres such as poetry exhibit distinctive linguistic characteristics compared to more conventional narrative forms like plays and novels. Novels, in turn, are inherently more narrative-driven than either drama or poetry, and thus possess features that further distinguish them. Our classification approach builds upon these core linguistic and structural differences. We focus on three main features to embed during training.

A \textit{syntax tree} represents the syntactic structure of a sentence. Initially introduced by Noam Chomsky in his seminal work \textit{Syntactic Structures} \citep{chomsky2002syntactic}, it has since become a foundational concept in modern linguistics \citep{Newmeyer_2023}. We incorporate syntactic information by providing both the \textit{syntax tree depth} and the \textit{depth-to-length ratio} (i.e., tree depth divided by sentence length).

We examine a representative lexical feature: \textit{metaphor count}. This is particularly useful for capturing stylistic and semantic distinctions. Metaphors reflect the density of figurative language, which tends to be more prevalent in expressive genres like poetry. We employ a metaphor detection model \citep{wachowiak2022drum} to count the number of metaphors per sentence, thereby gaining insight into the lexical choices that characterize each genre.

Phonetic features are among the most significant markers for distinguishing poetry from other literary genres \citep{vzirmunskij2016introduction}. These are most commonly realized through \textit{metre}—a system by which poets organize stress patterns, scansion, and metrical feet. The presence of metre is widely acknowledged as a key factor differentiating poetry, verse drama, and poetic prose from other genres. To evaluate whether metre is present in a sentence, we use \texttt{Poetry Tools}, a well-established Python library for analyzing metrical patterns.

Our task is defined as follows: for each binary classification instance, given an input \textit{I} and an encoder-only model \(E_\phi(\cdot)\), we aim to predict the label \(\hat{y} \in \{0,1\}\) by computing:
\begin{equation}
\hat{y} = \mathbf{1}(f_\theta(E_\phi(I)) > 0.5),
\end{equation}
where \(f_\theta(\cdot)\) denotes the classification head built on top of the encoder, which may consist of a linear layer, a multi-layer perceptron, or any other parameterized transformation. 

Originally, the input is defined as \(I = S\), where \(S\) is the raw sentence. To investigate whether linguistic features can enhance classification performance, we construct enriched inputs by appending additional feature representations to the sentence. Specifically, we define:
\begin{equation}
I = S \oplus F,
\end{equation}
where \(F\) denotes a feature vector encoding one of the linguistic features, and \(\oplus\) represents concatenation in the input space. We experiment with the following features:

\begin{itemize}
  \item \textbf{Syntax Tree Information:} represented as a tuple \((d, r)\), where \(d\) is the syntax tree depth and \(r = \frac{d}{|S|}\) is the depth-to-length ratio.
  \item \textbf{Metaphor Count:} a scalar feature \(m\) extracted by a pre-trained metaphor detection model, indicating the number of metaphorical tokens in \(S\).
  \item \textbf{Metre Pattern:} To extract phonetic rhythm, we compute a binary stress pattern vector:
    \begin{equation}
    \textit{MP}(S) = [m_1, m_2, \dots, m_n], \quad m_i \in \{0,1\},
    \end{equation}
    where \( m_i = 1 \) indicates a stressed syllable and \( m_i = 0 \) indicates an unstressed syllable for the \(i\)-th syllable in sentence \(S\).
\end{itemize}

\colorlet{improve}{green!15}
\colorlet{decline}{red!15}

Each feature is appended to the sentence embedding before being passed into the encoder. By comparing model performance on the baseline input \(I = S\) and enriched versions \(I = S \oplus F\), we aim to assess whether these linguistic features improve genre classification performance across different language-model pairs. 

\section{Experiments}

\begin{table*}[t]
  \renewcommand{\arraystretch}{1.05} 
  \begin{center}{
  \caption{F1 score statistics by genre, model and language (Percentage). We also compute the average result across all languages.}
  \vspace{-0.7em}
  {\footnotesize \textbf{Note}: P-Poetry, N-Novel, D-Drama.}
  \vspace{0.15em}\label{tab:F1 Scores}

  \newcolumntype{C}{>{\arraybackslash\smash}c}
  \resizebox{0.93\textwidth}{!}{\begin{tabular}{CCCCCCCCCCCCCCC}
    \toprule
    \multirow{2}{*}{\raisebox{-0.3em}{\textbf{Language}}}  & \multicolumn{3}{c}{\textbf{EN}} & \multicolumn{3}{c}{\textbf{FR}} & \multicolumn{3}{c}{\textbf{DE}} \\ 
    \cmidrule(lr){2-4} \cmidrule(lr){5-7} \cmidrule(lr){8-10}
    & \footnotesize P / N & \footnotesize P / D & \footnotesize N / D & \footnotesize P / N & \footnotesize P / D & \footnotesize N / D & \footnotesize P / N & \footnotesize P / D & \footnotesize N / D
    \\ \midrule
    BERT & 0.97 / 0.97 & 0.97 / 0.89 & 0.90 / 0.67 & 0.50 / 0.40 & 0.75 / 0.71 & 0.76 / 0.67 & 0.73 / 0.78 & 0.72 / 0.70 & 0.78 / 0.75
    \\ 
  DistilBERT & 0.97 / 0.97 & 0.97 / 0.89 & 0.89 / 0.68 & 0.76 / 0.59 & 0.75 / 0.71 & 0.75 / 0.71 & 0.74 / 0.72 & 0.75 / 0.68 & 0.77 / 0.72\\ 
    RoBERTa & 0.94 / 0.92 & 0.93 / 0.73 & 0.90 / 0.68 & 0.78 / 0.82 & 0.76 / 0.71 & 0.76 / 0.71 & 0.60 / 0.70 & 0.66 / 0.63 & 0.78 / 0.76 \\ 
    Metaphor RoBERTa & 0.94 / 0.92 & 0.92 / 0.69 & 0.89 / 0.67 & 0.77 / 0.82 & 0.75 / 0.75 & 0.79 / 0.68 & 0.78 / 0.82 & 0.79 / 0.75 & 0.80 / 0.79
    \\ \bottomrule
    \end{tabular}}

    \vspace{0.5em}

      \resizebox{0.93\textwidth}{!}{\begin{tabular}{CCCCCCCCCC}
      \toprule
      \multirow{2}{*}{\raisebox{-0.3em}{\textbf{Language}}}  & \multicolumn{3}{c}{\textbf{ES}} & \multicolumn{3}{c}{\textbf{IT}} & \multicolumn{3}{c}{\textbf{PT}}\\
      \cmidrule(lr){2-4} \cmidrule(lr){5-7} \cmidrule(lr){8-10} 
      & \footnotesize P / N & \footnotesize P / D & \footnotesize N / D & \footnotesize P / N & \footnotesize P / D & \footnotesize N / D & \footnotesize P / N & \footnotesize P / D & \footnotesize N / D \\
      \midrule
      BERT & 0.77 / 0.78 & 0.74 / 0.70 & 0.77 / 0.70 & 0.78 / 0.81 & 0.81 / 0.75 & 0.80 / 0.69 & 0.80 / 0.84 & 0.75 / 0.64 & 0.75 / 0.67 \\
      DistilBERT & 0.77 / 0.78 & 0.75 / 0.70 & 0.75 / 0.69 & 0.76 / 0.81 & 0.82 / 0.76 & 0.80 / 0.70 & 0.79 / 0.83 & 0.73 / 0.63 & 0.76 / 0.67 \\
      RoBERTa & 0.81 / 0.83 & 0.71 / 0.63 & 0.79 / 0.69 & 0.83 / 0.86 & 0.82 / 0.79 & 0.82 / 0.71 & 0.87 / 0.89 & 0.75 / 0.67 & 0.81 / 0.73 \\
      Metaphor RoBERTa & 0.80 / 0.82 & 0.78 / 0.77 & 0.82 / 0.71 & 0.82 / 0.85 & 0.80 / 0.78 & 0.82 / 0.72 & 0.86 / 0.89 & 0.77 / 0.68 & 0.81 / 0.72 \\
      \bottomrule
      \end{tabular}}}\end{center}
    \end{table*}

    \begin{table}[]
        \hspace{0.6em}
        \vspace{-2.4em}
        \begin{center}
        {\centering
        \newcolumntype{C}{>{\arraybackslash\smash}c}
        \resizebox{0.41\textwidth}{!}{\begin{tabular}{CCCC}
          \toprule
          \multirow{2}{*}{\raisebox{-0.3em}{\textbf{Language}}}  & \multicolumn{3}{c}{\textbf{Average}}  \\ 
          \cmidrule(lr){2-4} 
          & \footnotesize P / N & \footnotesize P / D & \footnotesize N / D
          \\ \midrule
          BERT & 0.76 / 0.76 & 0.79 / 0.73 &0.79 / 0.69 \\ 
          DistilBERT & 0.80 / 0.78 & 0.79 / 0.73 & 0.79 / 0.69 \\ 
          RoBERTa & 0.81 / 0.84 & 0.77 / 0.69 &0.81 / 0.71 \\ 
          Metaphor RoBERTa & 0.83 / 0.85 & 0.80 / 0.74 & 0.82 / 0.71 \\ \bottomrule
        \end{tabular}}}\end{center}
        \vspace{-0.8em}
        \begin{flushright}
            {\scriptsize
            }\rule{15pt}{0pt}
        \end{flushright}
        \vspace{-1em}
    \end{table}

In this section, we present the details of our experimental setup, report the results obtained, and provide a discussion of our findings.

\subsection{Experiment Settings}
\paragraph{Baseline} We fine-tune several pre-trained BERT-based models for genre classification tasks. As baseline models, we selected BERT~\citep{devlin2019bert}, DistilBERT~\citep{sanh2019distilbert}, RoBERTa~\citep{liu2019roberta}, and Metaphor-RoBERTa~\citep{wachowiak2022drum}. BERT and RoBERTa are widely recognized as robust encoder-only architectures and serve as standard baselines across numerous NLP tasks. DistilBERT, a lighter and faster variant of BERT, retains most of its language understanding capabilities, making it especially suitable for assessing how efficiently smaller models can learn linguistic features. Additionally, we incorporate Metaphor-RoBERTa, a model pre-trained specifically to detect metaphor usage, to investigate whether metaphor-aware pre-training enhances genre classification performance and impacts the effectiveness of our feature-enriched fine-tuning approach. 

\paragraph{Dataset and Training} We trained each model on the dataset described in Data Creation\ref{sec:dataset}, using a traditional supervised fine-tuning paradigm. For every binary genre classification task (e.g., poetry vs.~novel), we input each sentence—or feature-augmented sentence—into the model, and optimize the classification loss based on the ground truth labels. We repeated this process for each of the three binary classification tasks across six languages and four models, both with and without additional linguistic features (syntax tree depth, metaphor count, and metre pattern).

\paragraph{Evaluation} Finally, we compute the F1 scores for both genres in each binary classification task, yielding an F1 score pair \((F1_x, F1_y)\) for every combination \((t, l, m)\), where 
\(t \in T = \{t_1, t_2, \ldots, t_n\}\) represents the task set, 
\(l \in L = \{l_1, l_2, \ldots, l_p\}\) denotes the language set, and 
\(m \in M = \{m_1, m_2, \ldots, m_q\}\) is the model set. 
The F1 score for each class is calculated as:

\begin{align}
F1 = 2 \cdot \frac{\text{Precision} \cdot \text{Recall}}{\text{Precision} + \text{Recall}},
\end{align}

where Precision and Recall are defined in the standard way based on true positives, false positives, and false negatives.

To estimate the overall effectiveness of each method across languages, we compute the macro-average of the F1 scores for each task--model pair across all languages:

\begin{align}
\overline{F1}_{t,m} = \frac{1}{|L|} \sum_{l \in L} \frac{F1_x^{(t,l,m)} + F1_y^{(t,l,m)}}{2}.
\end{align}

This metric provides a balanced view of model performance by averaging scores for both genres and smoothing across language variations.

\subsection{Results}
\vspace{0.8em}

Table~\ref{tab:F1 Scores} presents the F1 score pairs obtained for each binary classification task across six languages (EN, FR, DE, ES, IT, PT) and four baseline models (BERT, DistilBERT, RoBERTa, and Metaphor-RoBERTa). Each cell reports the F1 scores for the two genres involved in a task.

Table~\ref{tab:optimization_comparison} summarizes the changes in F1 scores when incorporating three distinct linguistic features into the baseline models across the three genre classification tasks and across the languages. The detailed results are in Appendix A.

\paragraph{Results Across Languages.}\label{sec:baseline-results}
Across all languages, English and Portuguese consistently yield the highest performance across tasks and models, whereas French shows the greatest variability and relatively lower scores—especially with BERT and DistilBERT. We believe several factors contribute to this pattern:

(1) Pretraining bias: Most models are pretrained predominantly on English corpora, giving them stronger genre discrimination ability in English. They likely encounter frequent syntactic and lexical patterns in English during pretraining, making classification tasks especially easier.

\begin{figure*}[h!]
\centering

\subfigure[poetry vs. novel]{
\includegraphics[width=0.30\linewidth]{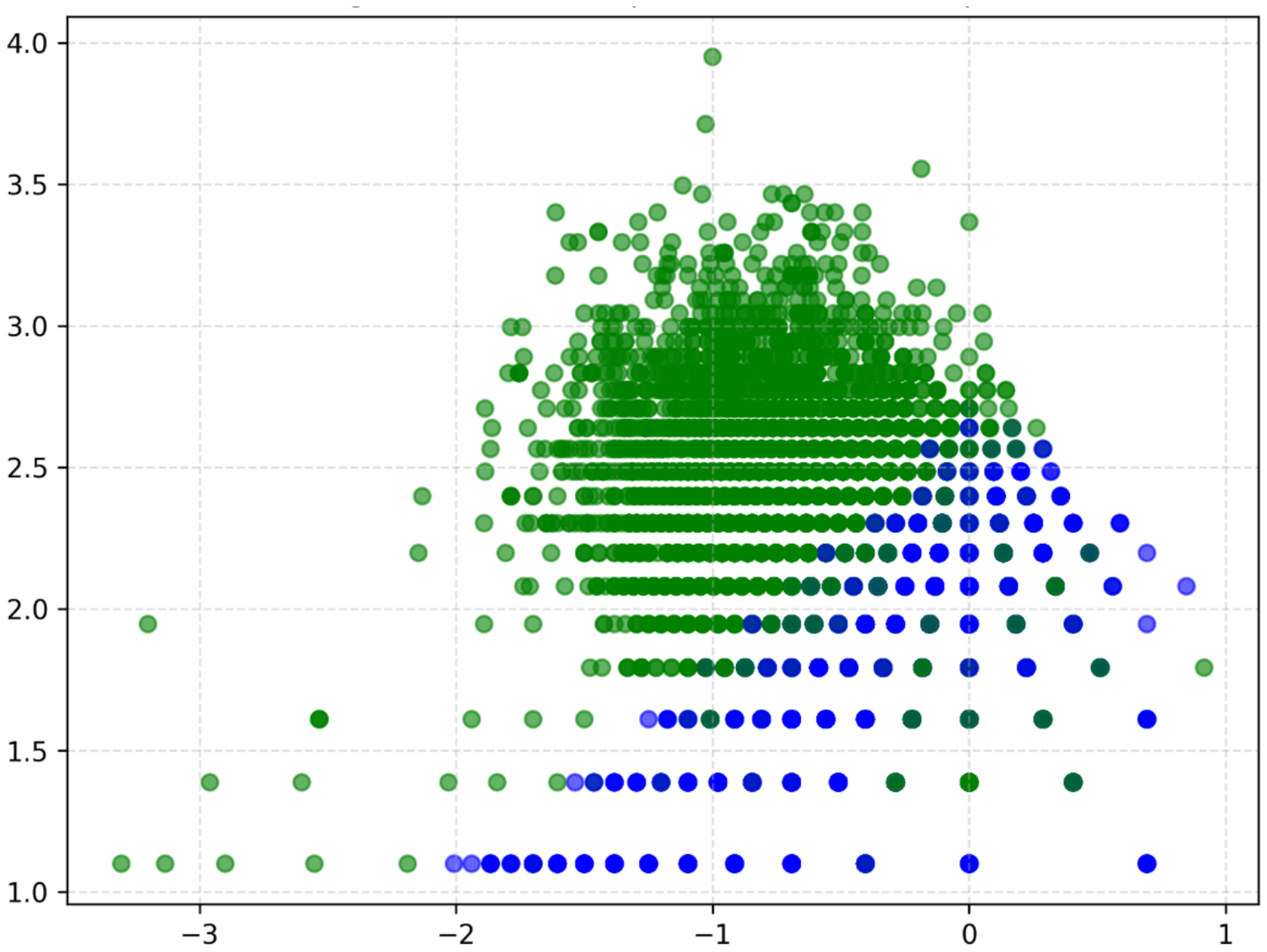}
}
\subfigure[poetry vs. novel (French)]{
\includegraphics[width=0.30\linewidth]{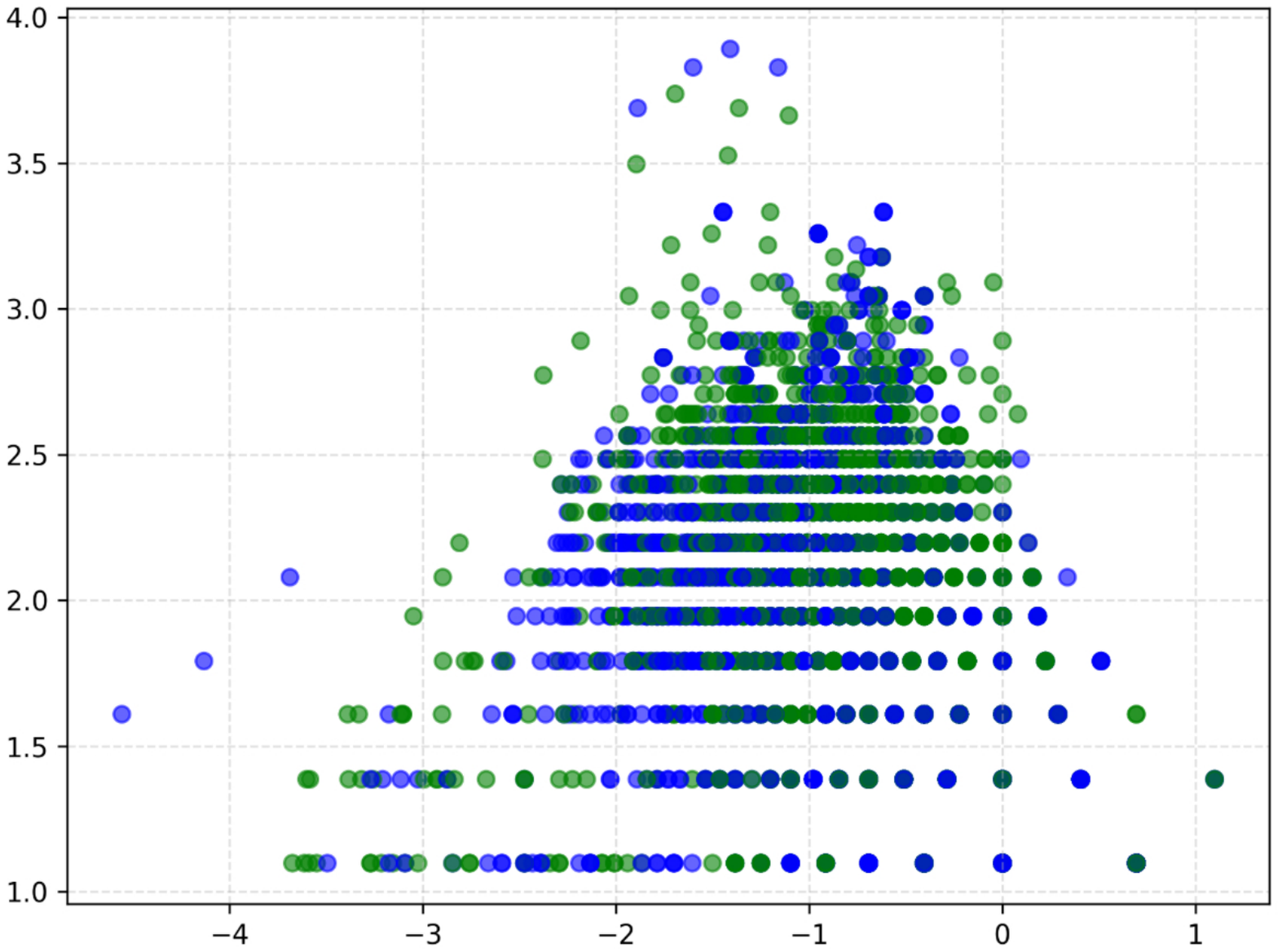}
}
\subfigure[novel vs. drama]{
\includegraphics[width=0.28\linewidth]{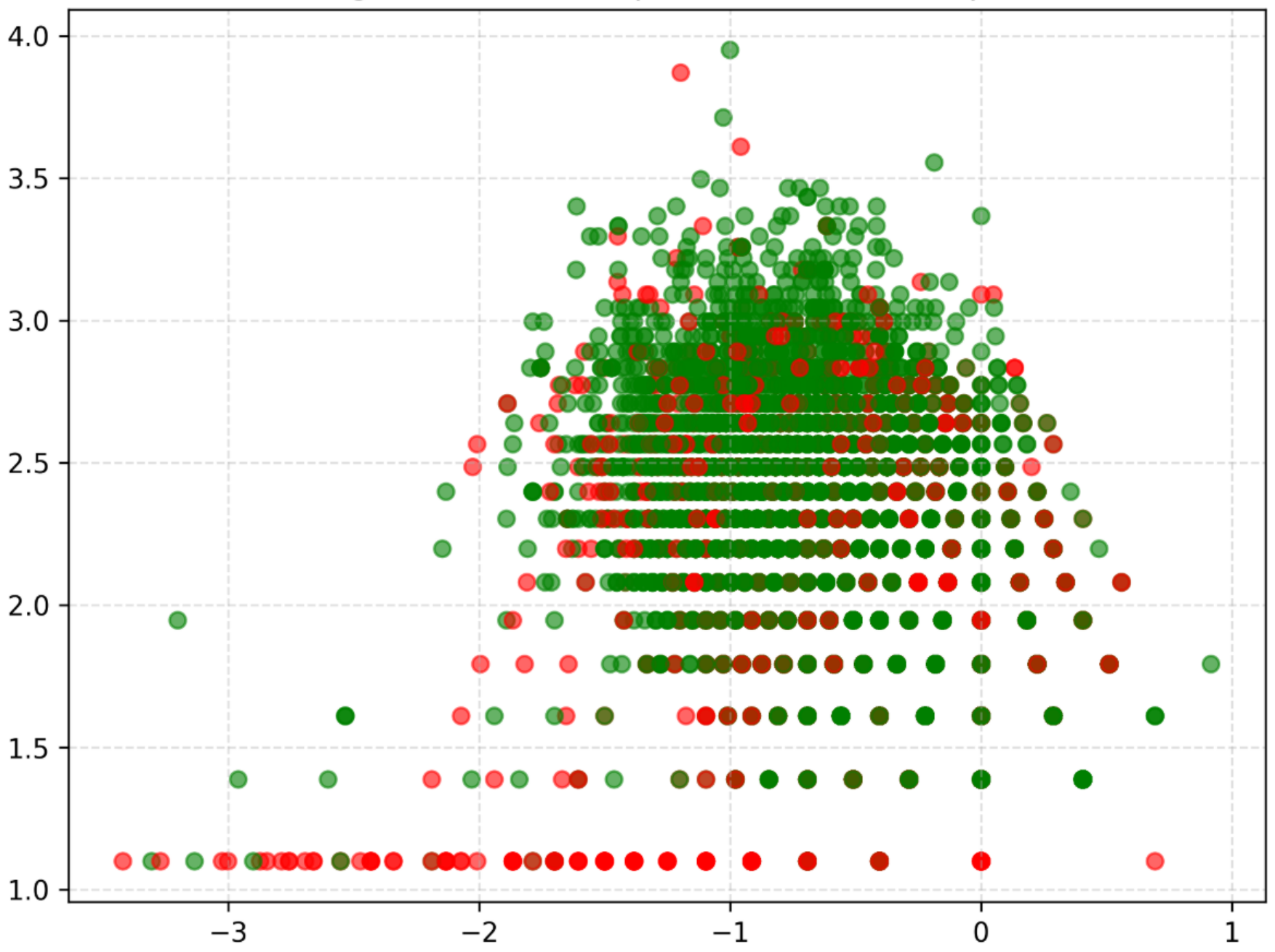}
}
\caption{Syntactic tree analysis. The x-axis represents $\log(\text{depth\_ratio})$, while the y-axis represents $\log(\text{tree\_depth} + 1)$. Green dots indicate novels, blue denote poetry, and red represent drama. Subfigure (a) shows the plot for Poetry vs. Novel in English, which is clearly linearly separable. Subfigure (b) presents the Poetry vs. Novel contrast in French, revealing a more complex distribution. Subfigure (c) displays the Novel vs. Drama set in English, which also exhibits significant overlap and is difficult to separate.}
\label{fig:syntax-comparison}
\end{figure*}

(2) Linguistic proximity between genres: In French, poetry and novels exhibit syntactic and stylistic structures that are more similar than in English. As shown in Figure~\ref{fig:syntax-comparison}, French poetry and novel texts are less separable based on syntactic tree depth and depth-to-length ratio compared to English, making differentiation more challenging.

\paragraph{Baseline Results Across Tasks.}
Genre pair differences are also notable. The \textit{Poetry vs. Novel} task is consistently the easiest across all models and languages, with average F1 scores generally exceeding 0.80. By contrast, \textit{Novel vs. Drama} is the most challenging, often yielding F1 scores below 0.75, likely due to overlapping narrative structures and lexical similarities between novels and dramatic texts. Poetry tends to employ distinctive linguistic features that are less common in prose forms. By contrast, both novels and dramas belong broadly to prose traditions: novels typically feature extended narrative, descriptive exposition, and inner monologue, whereas drama is conveyed primarily through dialogue and stage directions, but both share similar syntactic and lexical patterns. Consequently, it's easier for models to distinguish poetry from prose, while novel and drama are more linguistically proximate and thus harder to differentiate computationally.

\begin{figure*}[h!]
\centering

\subfigure[poetry vs. novel]{
\includegraphics[width=0.30\linewidth]{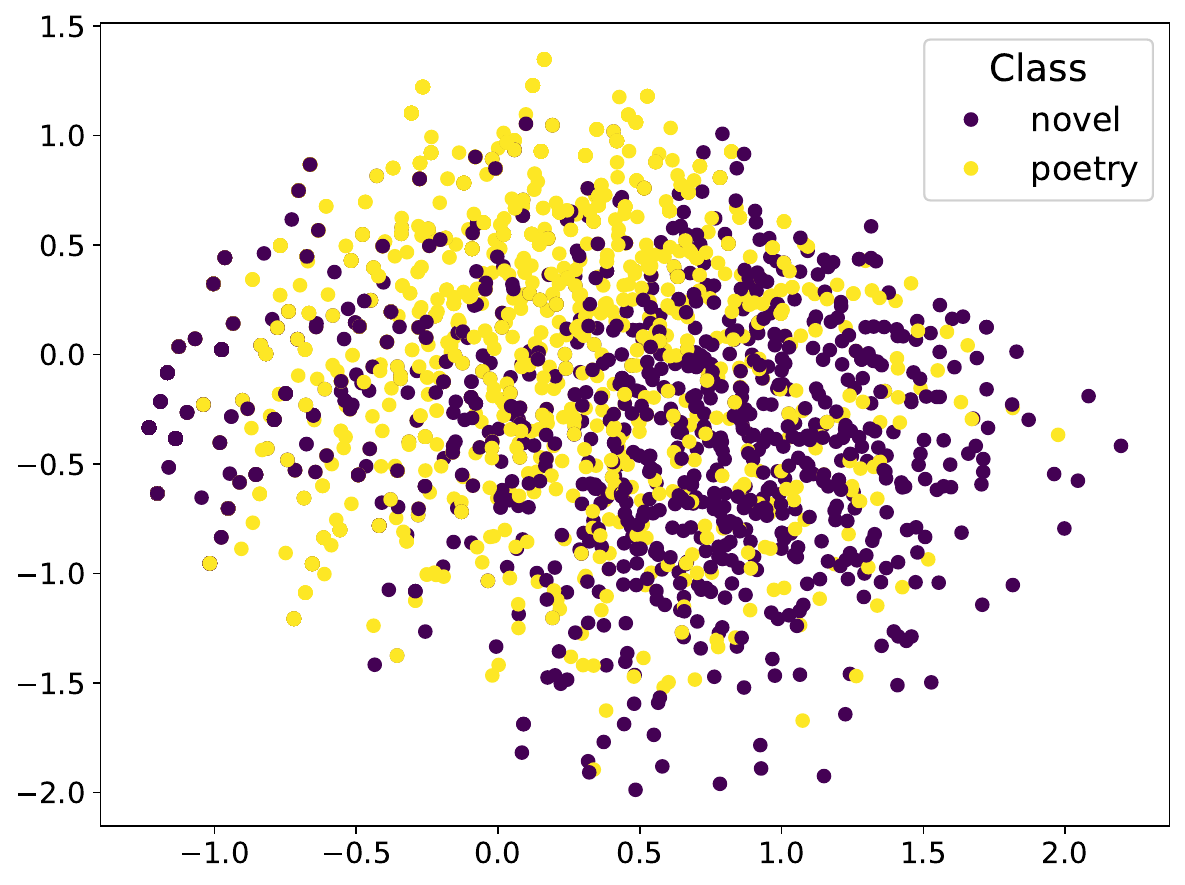}
}
\subfigure[drama vs. novel]{
\includegraphics[width=0.30\linewidth]{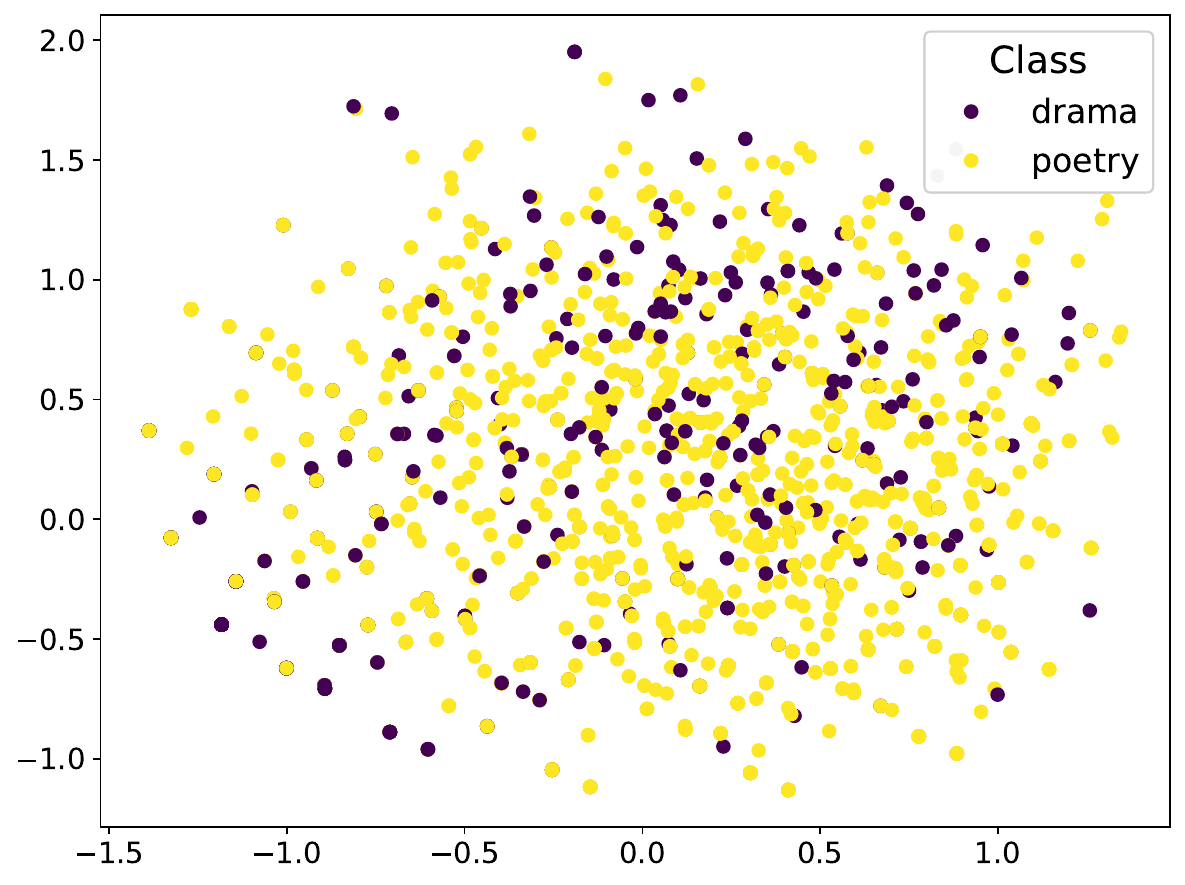}
}
\subfigure[drama vs. poetry]{
\includegraphics[width=0.30\linewidth]{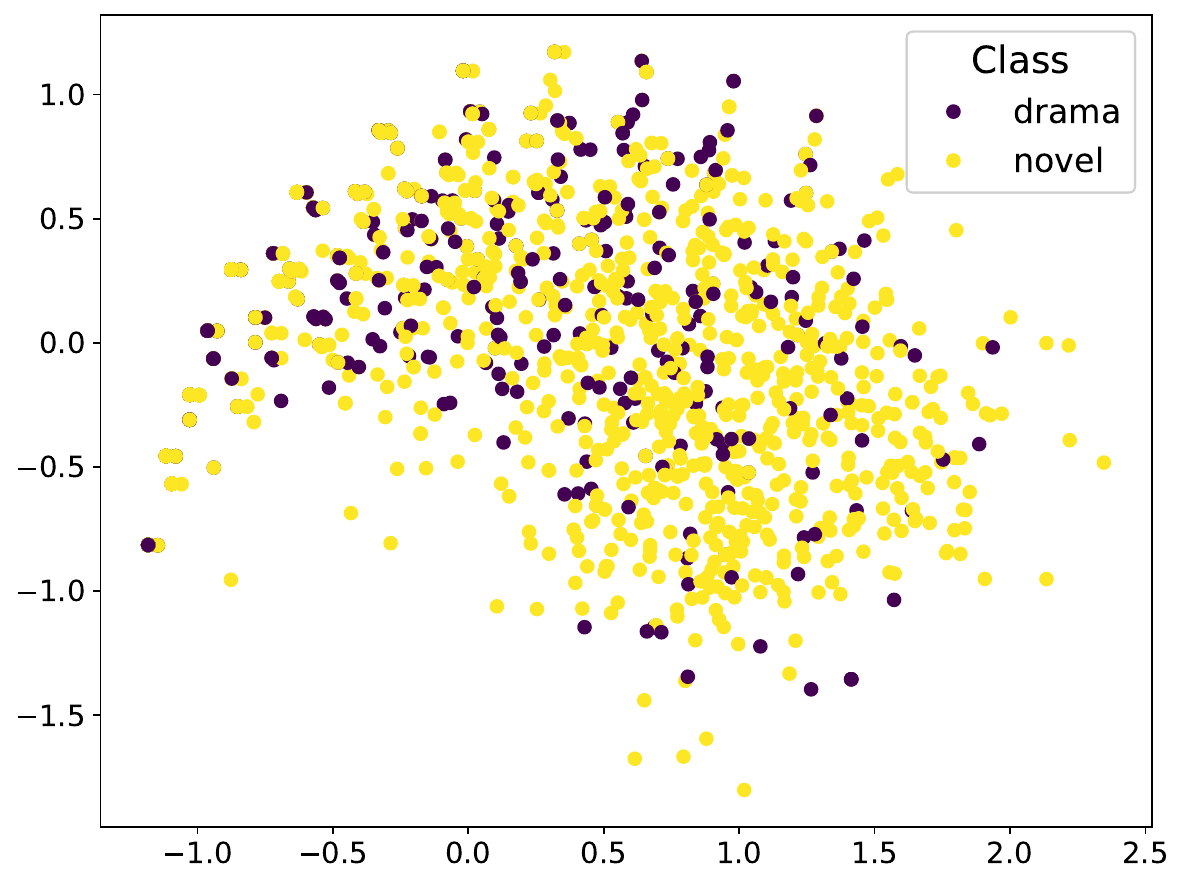}
}
\caption{Metre pattern analysis. We extract metre patterns from the raw texts and represent them as binary feature vectors, where each bit corresponds to a rhythmic unit. These vectors are then padded to uniform length and projected into a two-dimensional latent space using Principal Component Analysis (PCA). }
\label{fig:syntax-comparison-phonetic}
\end{figure*}

For the Novel vs. Drama task, all training methods fail to outperform the baseline, which is understandable given that these two genres often share similar narrative structures and linguistic features, making them inherently more difficult to distinguish. Beyond this, we observe varying degrees of improvement on the other two tasks across the four models.

Overall, syntax tree information and metaphor count do not appear to significantly enhance the models’ ability to differentiate between genres. Apart from negligible changes, only two configurations show notable improvements: the Poetry vs. Novel task with the BERT model, and the Poetry vs. Drama task with the RoBERTa model. A more detailed analysis will be provided in Section~\ref{sec:analysis}.

In contrast, the metre pattern feature proves to be the most robust and consistently beneficial across tasks and models. It delivers steady F1 score gains in both the Poetry vs. Novel and Poetry vs. Drama tasks, often outperforming other features by a margin of 2–7\%. These results underscore the importance of prosodic and rhythmic cues as fundamental signals of genre—particularly in poetry, where metrical structure is deeply embedded in the form.

\paragraph{Baseline Results Across Models.}
Across all languages and tasks, Metaphor-RoBERTa consistently demonstrates strong performance. It achieves the highest average F1 scores in both the \textit{Poetry vs. Novel} (0.83, 0.85) and \textit{Poetry vs. Drama} (0.80, 0.74) settings, and ties for the best result in \textit{Novel vs. Drama} (0.82, 0.71). This suggests that metaphor-aware pretraining may help capture stylistic distinctions between literary genres, particularly in tasks involving poetry.

RoBERTa also performs robustly, particularly on \textit{Poetry vs. Novel} tasks (0.81, 0.84), slightly trailing Metaphor-RoBERTa. Its advantage is especially pronounced in Romance languages such as Italian and Portuguese. DistilBERT, despite its smaller size, maintains competitive performance throughout. Notably, it outperforms BERT in many cases, including the French \textit{Poetry vs. Novel} task (0.76, 0.59 vs. 0.50, 0.40), demonstrating that reduced model capacity does not significantly compromise genre classification when fine-tuned properly.

BERT, while generally stable, underperforms in low-resource language scenarios (e.g., FR, DE), especially in the \textit{Poetry vs. Novel} task. This may indicate its relatively weaker adaptability to genre-specific linguistic variation in these languages.

Model-wise, BERT benefits the most across all three linguistic features, likely due to its larger capacity to leverage explicit signals. DistilBERT shows improvements primarily with metre patterns, possibly compensating for its reduced model size with clearer prosodic cues. RoBERTa and Metaphor-RoBERTa demonstrate mixed results, with metaphor-based fine-tuning occasionally lessening the positive impact of added linguistic features, possibly due to feature redundancy or overfitting.

\vspace{0.8em}

\begin{table*}
\centering
\caption{Comparative Analysis of Feature Optimization Methods (F1 Score Changes)}
\vspace{0.em}
\label{tab:optimization_comparison}
\footnotesize
\centering
\newcolumntype{C}[1]{>{\centering\arraybackslash\smash}p{#1}}
\resizebox{0.70\textwidth}{!}{
    \renewcommand{\arraystretch}{0.54} 
    \setlength{\extrarowheight}{1.8pt} 
        \begin{tabular}{lc|C{0.95cm} C{0.95cm}|C{0.95cm} C{0.95cm}|C{0.95cm} C{0.95cm}|C{0.95cm} C{0.95cm}}
            \toprule
\multirow{2}{*}{{\textbf{Feature}}}
& \multirow{2}{*}{{\textbf{Set}}}
  & \multicolumn{2}{c}{\textbf{BERT}} 
  & \multicolumn{2}{c}{\textbf{DistilBERT}} 
  & \multicolumn{2}{c}{\textbf{RoBERTa}} 
  & \multicolumn{2}{c}{\makecell{\small\textbf{Metaphor}\\[-2pt]\small\textbf{RoBERTa}}} \\

            \midrule
            
            \rowcolor{gray!10}
            \multicolumn{10}{l}{\rule{0pt}{10pt}\textbf{Syntax Tree Depth}} \\
            \multirow{2}{*}{} & \multirow{2}{*}{Poetry+Novel}
              & \rule{0pt}{10pt}\cellcolor{improve}0.77 & \cellcolor{improve}0.81
              &0.77 & 0.80
              & 0.82 & 0.85
              & 0.83 & 0.86 \\
            & 
              & \cellcolor{improve}\color{green!30!black}\scriptsize +1\% & \cellcolor{improve}\color{green!30!black}\scriptsize +5\%
              & \color{red!70!black}\scriptsize -3\% & \color{green!30!black}\scriptsize +2\%
              & \color{green!30!black}\scriptsize +1\% & \color{green!30!black}\scriptsize +1\%
              & \scriptsize -- & \color{green!30!black}\scriptsize +1\% \\
            
              \multirow{2}{*}{} & \multirow{2}{*}{Poetry+Drama}
              & 0.79 & \rule{0pt}{10pt}0.72
              & 0.78 & 0.72
              & \cellcolor{improve}0.77 & \cellcolor{improve}0.72
              & \cellcolor{decline}0.79 & \cellcolor{decline}0.66 \\
            & 
              & \scriptsize -- & \color{red!70!black}\scriptsize -1\%
              & \color{red!70!black}\scriptsize -1\% & \color{red!70!black}\scriptsize -1\%
              & \cellcolor{improve}\scriptsize -- & \cellcolor{improve}\color{green!30!black}\scriptsize +3\%
              & \cellcolor{decline}\color{red!70!black}\scriptsize -1\% & \cellcolor{decline}\color{red!70!black}\scriptsize -8\% \\
            
              \multirow{2}{*}{} & \multirow{2}{*}{Novel+Drama}
              & 0.79 & 0.68
              & 0.78 & 0.69\rule{0pt}{10pt}
              & 0.81 & 0.71
              & \cellcolor{decline}0.80 & \cellcolor{decline}0.67 \\
            & 
              & \scriptsize -- & \color{red!70!black}\scriptsize -1\%
              & \color{red!70!black}\scriptsize -1\% & \scriptsize --
              & \scriptsize -- & \scriptsize --
              & \cellcolor{decline}\color{red!70!black}\scriptsize -2\% & \cellcolor{decline}\color{red!70!black}\scriptsize -4\% \\
            
            \midrule
            \rowcolor{gray!10}
            \multicolumn{10}{l}{\textbf{\rule{0pt}{10pt}Metaphor Count}} \\
            \multirow{2}{*}{} & \multirow{2}{*}{Poetry+Novel}
              & \cellcolor{improve}0.78 & \cellcolor{improve}0.80
              & 0.78 & 0.80\rule{0pt}{10pt}
              & 0.82 & 0.85
              & 0.84 & 0.86 \\
            & 
              & \cellcolor{improve}\color{green!30!black}\scriptsize +2\% & \cellcolor{improve}\color{green!30!black}\scriptsize +4\%
              & \scriptsize\color{red!70!black}\ -2\% & \color{green!30!black}\scriptsize +2\%
              & \color{green!30!black}\scriptsize +1\% & \color{green!30!black}\scriptsize +1\%
              & \color{green!30!black}\scriptsize +1\% & \color{green!30!black}\scriptsize +1\% \\
            
              \multirow{2}{*}{} & \multirow{2}{*}{Poetry+Drama}
              & 0.79 & 0.73\rule{0pt}{10pt}
              & \raisebox{-0ex}{0.78} & 0.73
              & \cellcolor{improve}0.77 & \cellcolor{improve}0.71
              & 0.80 & 0.76 \\
            & 
              & \scriptsize -- & \scriptsize --
              & \color{red!70!black}\raisebox{0ex}{
              \scriptsize -1\%} & \scriptsize --
              & \cellcolor{improve}\scriptsize -- & \cellcolor{improve}\color{green!30!black}\scriptsize +2\%
              & \color{red!70!black}\scriptsize -2\% & \color{green!30!black}\scriptsize +2\% \\
            
              \multirow{2}{*}{} & \multirow{2}{*}{Novel+Drama}
              & 0.78 & 0.69
              & 0.79 & 0.69
              & \cellcolor{decline}0.81 & \cellcolor{decline}0.69\rule{0pt}{10pt}
              & 0.81 & 0.70 \\
            & 
              & \color{red!70!black}\scriptsize -1\% & \scriptsize --
              & \scriptsize -- & \scriptsize --
              & \cellcolor{decline}\scriptsize -- & \cellcolor{decline}\color{red!70!black}\scriptsize -2\%
              & \color{red!70!black}\scriptsize -1\% & \color{red!70!black}\scriptsize -1\% \\
            
            \midrule
            \rowcolor{gray!10}
            \multicolumn{10}{l}{\rule{0pt}{10pt}\textbf{Metre Pattern}} \\
            \multirow{2}{*}{} & \multirow{2}{*}{Poetry+Novel}
              & \cellcolor{improve}0.80 & \cellcolor{improve}0.83\rule{0pt}{10pt}
              & \cellcolor{improve}0.79 & \cellcolor{improve}0.81
              & \cellcolor{decline}0.79 & \cellcolor{decline}0.84
              & \cellcolor{improve}0.83 & \cellcolor{improve}0.87 \\
            & 
              & \cellcolor{improve}\color{green!30!black}\scriptsize +4\% & \cellcolor{improve}\color{green!30!black}\scriptsize +7\%
              & \cellcolor{improve}\color{red!70!black}\scriptsize -1\% & \cellcolor{improve}\color{green!30!black}\scriptsize +3\%
              & \cellcolor{decline}\color{red!70!black}\scriptsize -2\% & \cellcolor{decline}\scriptsize --
              & \cellcolor{improve}\scriptsize --  & \cellcolor{improve}\color{green!30!black}\scriptsize +2\%\\
            
            \multirow{2}{*}{} & \multirow{2}{*}{Poetry+Drama}
              & 0.80 & 0.72\rule{0pt}{10pt}
              & 0.79 & 0.74
              & \cellcolor{improve}0.78 & \cellcolor{improve}0.73
              & \cellcolor{improve}0.81 & \cellcolor{improve}0.76 \\
            & 
              & \color{green!30!black}\scriptsize +1\% & \color{red!70!black}\scriptsize -1\%
              & \scriptsize -- & \color{green!30!black}\scriptsize +1\%
              & \cellcolor{improve}\color{green!30!black}\scriptsize +1\% & \cellcolor{improve}\color{green!30!black}\scriptsize +4\%
              & \cellcolor{improve}\color{green!30!black}\scriptsize +1\% & \cellcolor{improve}\color{green!30!black}\scriptsize +2\% \\
            
            \multirow{2}{*}{} & \multirow{2}{*}{Novel+Drama}
              & 0.79 & 0.69
              & 0.78 & 0.69
              & \cellcolor{decline}0.79 & \cellcolor{decline}0.66\rule{0pt}{10pt}
              & \cellcolor{decline}0.77 & \cellcolor{decline}0.66 \\
            & 
              & \scriptsize -- & \scriptsize --
              & \color{red!70!black}\scriptsize -1\% & \scriptsize --
              & \cellcolor{decline}\color{red!70!black}\scriptsize -2\% & \cellcolor{decline}\color{red!70!black}\scriptsize -5\%
              & \cellcolor{decline}\color{red!70!black}\smash{\scriptsize -5\%} & \cellcolor{decline}\color{red!70!black}\scriptsize -5\% \\
            
            \bottomrule
            \end{tabular}}
            \parbox{\textwidth}{
                \vspace{0.2em}\hspace{1.2em}
\scriptsize
\centering
\textbf{Color of notation:} {\color{green!30!black}{$\uparrow$} Improvement  }{\color{red!70!black}{$\downarrow$} Decline }\qquad
\textbf{Color of cell}: \colorbox{improve}{Improvement $\geq$ 0.02} | \colorbox{decline}{Decline $\geq$ 0.02} (non-negligible changes)
}
\end{table*}

\vspace{0.8em}

\subsection{Results Analysis} 
\label{sec:analysis}
In this section, we analyze how three linguistic features contribute to improving model performance on the classification task.

\paragraph{Syntax Tree Information.}
As shown in Table~\ref{tab:optimization_comparison}, \textit{Metre Pattern} achieved the most substantial performance improvement among the three linguistic features. The other two features yielded only slight gains, primarily on the \textit{Poetry vs. Novel} classification task, and showed limited effectiveness on the other two genre pairs. This can be explained from a linguistic perspective. Taking English as an example, Figure~\ref{fig:syntax-comparison} reveal that syntax tree depth and depth ratio exhibit clear linear separability between poetry and novel, whereas the distinction between drama and novel is far less apparent. Given the relatively distinct boundary between poetry and novel, it is likely that the model is already capable of capturing syntactic information directly from raw sentences, which may explain the modest gains observed when these syntactic features are added through our fine-tuning method. Additionally, we use the same metric to evaluate the French dataset. The results reveal that French poetry and novels are not easily separable based on syntactic tree information, which may also explain the limited improvement observed on the French data.

\paragraph{Metaphor Count.}
Meanwhile, we calculated the average number of metaphorical words per sentence in English texts across three genres: drama (1.38), novel (2.50), and poetry (1.39). The results show that drama and poetry share similar levels of metaphor usage, while novels tend to contain significantly more metaphorical expressions on average. However, since the overall differences are relatively subtle, this variation does not appear to strongly influence the model’s ability to distinguish between genres. Although it does not directly contribute to performance, this method plays a key role in controlling for the potential impact of increased token length on our tasks. The results suggest that linguistic features such as metre patterns contribute to performance improvements that go beyond the effects of token length, highlighting the intrinsic value of the features themselves.

\paragraph{Metre Pattern.}
Most importantly, the average F1 score has increased significantly when metre patterns are taken into consideration. This suggests that metre patterns encode subtle but meaningful information that improves genre classification performance, even if such patterns are not easily separable in low-dimensional space. As shown in Figure~\ref{fig:syntax-comparison-phonetic}, the PCA projection of the binary metre vectors does not exhibit clear clustering by class, indicating that the informative features may be distributed across multiple dimensions and not linearly separable. This reinforces the idea that metre patterns contribute in a complex, high-dimensional manner that benefits classification models, even if they are not directly interpretable via 2D visualization. The improved F1 scores highlight that models can detect and leverage these latent rhythmic signals effectively, even when they are not visually obvious.

\section{Limitation}
While our study introduces a linguistically enriched, multilingual dataset and provides novel insights into literary genre classification with LLMs, several limitations remain. First, the dataset is still constrained by the availability and biases of Project Gutenberg texts. These works tend to overrepresent canonical literature from specific historical periods and underrepresent contemporary, non-Western, or marginalized voices, which may limit the generalizability of our findings. Second, the extraction of linguistic features such as metaphor and phonetic regularity relies on heuristic or proxy-based methods, which may miss nuanced or culturally specific expressions. Future work should explore the inclusion of more diverse corpora and refine feature extraction with neural or hybrid methods to support interpretability across typologically diverse languages.

\section{Conclusion}
This study presents a novel multilingual dataset for literary genre classification, enhanced with linguistically informed features such as syntactic depth, metaphor frequency, and metrical patterns. By integrating these cues into large language model pipelines, we show that genre distinctions can be computationally captured through measurable linguistic signals. Our results indicate that these features improve genre discrimination across languages, offering both performance gains and interpretability. This work lays the groundwork for future research at the intersection of linguistics, literary analysis, and AI, supporting cross-cultural stylistic inquiry and deeper engagement with the structural and figurative dimensions of text. The future direction of this research involves expanding the dataset to cover more languages and literary traditions, enabling broader cross-cultural comparisons, and applying the method to broader and more complex tasks.

\newpage

\bibliography{aaai2026}

\setlength{\leftmargini}{20pt}
\makeatletter\def\@listi{\leftmargin\leftmargini \topsep .5em \parsep .5em \itemsep .5em}
\def\@listii{\leftmargin\leftmarginii \labelwidth\leftmarginii \advance\labelwidth-\labelsep \topsep .4em \parsep .4em \itemsep .4em}
\def\@listiii{\leftmargin\leftmarginiii \labelwidth\leftmarginiii \advance\labelwidth-\labelsep \topsep .4em \parsep .4em \itemsep .4em}\makeatother

\setcounter{secnumdepth}{0}
\renewcommand\thesubsection{\arabic{subsection}}
\renewcommand\labelenumi{\thesubsection.\arabic{enumi}}

\newcounter{checksubsection}
\newcounter{checkitem}[checksubsection]

\newcommand{\checksubsection}[1]{%
  \refstepcounter{checksubsection}%
  \paragraph{\arabic{checksubsection}. #1}%
  \setcounter{checkitem}{0}%
}

\newcommand{\checkitem}{%
  \refstepcounter{checkitem}%
  \item[\arabic{checksubsection}.\arabic{checkitem}.]%
}
\newcommand{\question}[2]{\normalcolor\checkitem #1 #2 \color{blue}}
\newcommand{\ifyespoints}[1]{\makebox[0pt][l]{\hspace{-15pt}\normalcolor #1}}

\section*{Reproducibility Checklist}

\vspace{1em}
\hrule
\vspace{1em}


\checksubsection{General Paper Structure}
\begin{itemize}

\question{Includes a conceptual outline and/or pseudocode description of AI methods introduced}{(yes/partial/no/NA)}
yes

\question{Clearly delineates statements that are opinions, hypothesis, and speculation from objective facts and results}{(yes/no)}
yes

\question{Provides well-marked pedagogical references for less-familiar readers to gain background necessary to replicate the paper}{(yes/no)}
yes

\end{itemize}
\checksubsection{Theoretical Contributions}
\begin{itemize}

\question{Does this paper make theoretical contributions?}{(yes/no)}
no

	\ifyespoints{\vspace{1.2em}If yes, please address the following points:}
        \begin{itemize}
	
	\question{All assumptions and restrictions are stated clearly and formally}{(yes/partial/no)}
	Type your response here

	\question{All novel claims are stated formally (e.g., in theorem statements)}{(yes/partial/no)}
	Type your response here

	\question{Proofs of all novel claims are included}{(yes/partial/no)}
	Type your response here

	\question{Proof sketches or intuitions are given for complex and/or novel results}{(yes/partial/no)}
	Type your response here

	\question{Appropriate citations to theoretical tools used are given}{(yes/partial/no)}
	Type your response here

	\question{All theoretical claims are demonstrated empirically to hold}{(yes/partial/no/NA)}
	Type your response here

	\question{All experimental code used to eliminate or disprove claims is included}{(yes/no/NA)}
	Type your response here
	
	\end{itemize}
\end{itemize}

\checksubsection{Dataset Usage}
\begin{itemize}

\question{Does this paper rely on one or more datasets?}{(yes/no)}
yes

\ifyespoints{If yes, please address the following points:}
\begin{itemize}

	\question{A motivation is given for why the experiments are conducted on the selected datasets}{(yes/partial/no/NA)}
	yes

	\question{All novel datasets introduced in this paper are included in a data appendix}{(yes/partial/no/NA)}
	no

	\question{All novel datasets introduced in this paper will be made publicly available upon publication of the paper with a license that allows free usage for research purposes}{(yes/partial/no/NA)}
	yes

	\question{All datasets drawn from the existing literature (potentially including authors' own previously published work) are accompanied by appropriate citations}{(yes/no/NA)}
	yes

	\question{All datasets drawn from the existing literature (potentially including authors' own previously published work) are publicly available}{(yes/partial/no/NA)}
	yes

	\question{All datasets that are not publicly available are described in detail, with explanation why publicly available alternatives are not scientifically satisficing}{(yes/partial/no/NA)}
	NA

\end{itemize}
\end{itemize}

\checksubsection{Computational Experiments}
\begin{itemize}

\question{Does this paper include computational experiments?}{(yes/no)}
yes

\ifyespoints{If yes, please address the following points:}
\begin{itemize}

	\question{This paper states the number and range of values tried per (hyper-) parameter during development of the paper, along with the criterion used for selecting the final parameter setting}{(yes/partial/no/NA)}
	no

	\question{Any code required for pre-processing data is included in the appendix}{(yes/partial/no)}
	no

	\question{All source code required for conducting and analyzing the experiments is included in a code appendix}{(yes/partial/no)}
	no

	\question{All source code required for conducting and analyzing the experiments will be made publicly available upon publication of the paper with a license that allows free usage for research purposes}{(yes/partial/no)}
	yes
        
	\question{All source code implementing new methods have comments detailing the implementation, with references to the paper where each step comes from}{(yes/partial/no)}
	no

	\question{If an algorithm depends on randomness, then the method used for setting seeds is described in a way sufficient to allow replication of results}{(yes/partial/no/NA)}
	NA

	\question{This paper specifies the computing infrastructure used for running experiments (hardware and software), including GPU/CPU models; amount of memory; operating system; names and versions of relevant software libraries and frameworks}{(yes/partial/no)}
	no

	\question{This paper formally describes evaluation metrics used and explains the motivation for choosing these metrics}{(yes/partial/no)}
	yes

	\question{This paper states the number of algorithm runs used to compute each reported result}{(yes/no)}
	no

	\question{Analysis of experiments goes beyond single-dimensional summaries of performance (e.g., average; median) to include measures of variation, confidence, or other distributional information}{(yes/no)}
	no

	\question{The significance of any improvement or decrease in performance is judged using appropriate statistical tests (e.g., Wilcoxon signed-rank)}{(yes/partial/no)}
	yes

	\question{This paper lists all final (hyper-)parameters used for each model/algorithm in the paper’s experiments}{(yes/partial/no/NA)}
	partial

\end{itemize}
\end{itemize}

\appendix
\onecolumn

\renewcommand{\thetable}{A\arabic{table}}
\renewcommand{\thefigure}{A\arabic{figure}}
\setcounter{table}{0}
\setcounter{figure}{0}

\section{A: Detailed Performance under Different Linguistic Features}
\label{detailedinfo}
Below, we present the detailed F1 scores obtained under our three linguistic feature–based training approaches. The experimental settings are kept consistent with those used in the baseline model to ensure fair comparison. The only difference lies in the incorporation of linguistic features—each model is enhanced by adding one of the following types of features: syntactic (e.g., syntax tree depth), phonological (e.g., metre patterns), or stylistic (e.g., metaphor usage). These features are intended to inject deeper linguistic knowledge into the model, thereby improving its ability to distinguish between genres. The results provide insight into the individual contributions of these features to classification performance.

\begin{table*}[h]
\centering
\caption{F1 scores by genre and language with \textbf{syntax tree depth} considered.}
\label{tab:syntax-depth}
\begin{tabular}{llcccc}
\toprule
\textbf{Language} & \textbf{Set} & \textbf{BERT} & \textbf{DistilBERT} & \textbf{RoBERTa} & \textbf{Metaphor RoBERTa} \\
\midrule
\multirow{3}{*}{EN} 
& Poetry + Novel     & 0.97+0.96 & 0.96+0.95 & 0.97+0.96 & 0.97+0.96 \\
& Poetry + Drama     & 0.97+0.88 & 0.96+0.86 & 0.97+0.89 & 0.96+0.87 \\
& Novel + Drama      & 0.90+0.68 & 0.90+0.67 & 0.90+0.64 & 0.89+0.62 \\
\cmidrule(lr){1-6}
\multirow{3}{*}{FR} 
& Poetry + Novel     & 0.64+0.74 & 0.66+0.71 & 0.78+0.82 & 0.75+0.81 \\
& Poetry + Drama     & 0.72+0.72 & 0.77+0.73 & 0.69+0.69 & 0.73+0.70 \\
& Novel + Drama      & 0.72+0.70 & 0.71+0.65 & 0.75+0.74 & 0.76+0.70 \\
\cmidrule(lr){1-6}
\multirow{3}{*}{DE} 
& Poetry + Novel     & 0.73+0.77 & 0.72+0.77 & 0.74+0.76 & 0.80+0.83 \\
& Poetry + Drama     & 0.76+0.67 & 0.74+0.69 & 0.75+0.70 & 0.76+0.74 \\
& Novel + Drama      & 0.78+0.75 & 0.78+0.72 & 0.81+0.74 & 0.80+0.74 \\
\cmidrule(lr){1-6}
\multirow{3}{*}{ES} 
& Poetry + Novel     & 0.75+0.77 & 0.77+0.76 & 0.79+0.81 & 0.78+0.81 \\
& Poetry + Drama     & 0.76+0.69 & 0.70+0.69 & 0.72+0.71 & 0.69+0.22 \\
& Novel + Drama      & 0.74+0.69 & 0.74+0.71 & 0.76+0.67 & 0.79+0.64 \\
\cmidrule(lr){1-6}
\multirow{3}{*}{IT} 
& Poetry + Novel     & 0.76+0.79 & 0.74+0.80 & 0.77+0.84 & 0.82+0.85 \\
& Poetry + Drama     & 0.78+0.72 & 0.79+0.73 & 0.75+0.71 & 0.80+0.77 \\
& Novel + Drama      & 0.79+0.69 & 0.80+0.72 & 0.83+0.74 & 0.81+0.71 \\
\cmidrule(lr){1-6}
\multirow{3}{*}{PT} 
& Poetry + Novel     & 0.76+0.81 & 0.76+0.81 & 0.86+0.89 & 0.86+0.89 \\
& Poetry + Drama     & 0.73+0.62 & 0.74+0.65 & 0.71+0.64 & 0.77+0.67 \\
& Novel + Drama      & 0.78+0.60 & 0.76+0.68 & 0.80+0.70 & 0.73+0.62 \\
\cmidrule(lr){1-6}
\multirow{3}{*}{Average} 
& Poetry + Novel     & 0.77+0.81 & 0.77+0.80 & 0.82+0.85 & 0.83+0.86 \\
& Poetry + Drama     & 0.79+0.72 & 0.78+0.72 & 0.77+0.72 & 0.79+0.66 \\
& Novel + Drama      & 0.79+0.68 & 0.78+0.69 & 0.81+0.71 & 0.80+0.67 \\
\bottomrule
\end{tabular}
\end{table*}

\begin{table*}[h]
\centering
\caption{F1 scores by genre and language with \textbf{metaphor count} considered.}
\label{tab:metaphor-count}
\begin{tabular}{llcccc}
\toprule
\textbf{Language} & \textbf{Set} & \textbf{BERT} & \textbf{DistilBERT} & \textbf{RoBERTa} & \textbf{Metaphor RoBERTa} \\
\midrule

\multirow{3}{*}{EN} 
& Poetry + Novel     & 0.97+0.96 & 0.97+0.97 & 0.97+0.96 & 0.96+0.96 \\
& Poetry + Drama     & 0.97+0.89 & 0.97+0.90 & 0.97+0.88 & 0.97+0.89 \\
& Novel + Drama      & 0.91+0.69 & 0.91+0.68 & 0.89+0.59 & 0.90+0.64 \\
\cmidrule(lr){1-6}

\multirow{3}{*}{FR} 
& Poetry + Novel     & 0.66+0.73 & 0.70+0.67 & 0.74+0.82 & 0.79+0.81 \\
& Poetry + Drama     & 0.75+0.73 & 0.71+0.73 & 0.69+0.70 & 0.64+0.72 \\
& Novel + Drama      & 0.71+0.72 & 0.72+0.68 & 0.76+0.70 & 0.74+0.68 \\
\cmidrule(lr){1-6}

\multirow{3}{*}{DE} 
& Poetry + Novel     & 0.73+0.75 & 0.73+0.78 & 0.80+0.84 & 0.80+0.84 \\
& Poetry + Drama     & 0.76+0.69 & 0.72+0.71 & 0.73+0.70 & 0.76+0.73 \\
& Novel + Drama      & 0.78+0.73 & 0.78+0.73 & 0.81+0.76 & 0.79+0.76 \\
\cmidrule(lr){1-6}

\multirow{3}{*}{ES} 
& Poetry + Novel     & 0.76+0.74 & 0.79+0.77 & 0.79+0.76 & 0.81+0.82 \\
& Poetry + Drama     & 0.74+0.73 & 0.73+0.72 & 0.74+0.71 & 0.77+0.75 \\
& Novel + Drama      & 0.71+0.68 & 0.77+0.67 & 0.80+0.70 & 0.81+0.68 \\
\cmidrule(lr){1-6}

\multirow{3}{*}{IT} 
& Poetry + Novel     & 0.77+0.79 & 0.76+0.81 & 0.75+0.83 & 0.79+0.85 \\
& Poetry + Drama     & 0.78+0.73 & 0.79+0.73 & 0.75+0.70 & 0.80+0.79 \\
& Novel + Drama      & 0.79+0.69 & 0.81+0.69 & 0.83+0.73 & 0.81+0.72 \\
\cmidrule(lr){1-6}

\multirow{3}{*}{PT} 
& Poetry + Novel     & 0.79+0.82 & 0.75+0.80 & 0.86+0.89 & 0.87+0.89 \\
& Poetry + Drama     & 0.73+0.63 & 0.73+0.60 & 0.75+0.57 & 0.75+0.67 \\
& Novel + Drama      & 0.79+0.66 & 0.77+0.67 & 0.78+0.64 & 0.82+0.71 \\
\cmidrule(lr){1-6}

\multirow{3}{*}{Average} 
 & Poetry + Novel    & 0.78+0.80 & 0.78+0.80 & 0.82+0.85 & 0.84+0.86 \\
 & Poetry + Drama    & 0.79+0.73 & 0.78+0.73 & 0.77+0.71 & 0.78+0.76 \\
 & Novel + Drama     & 0.78+0.69 & 0.79+0.69 & 0.81+0.69 & 0.81+0.70 \\
\bottomrule

\end{tabular}
\end{table*}

\begin{table*}[h]
\centering
\caption{F1 scores by genre and language with \textbf{metre pattern} considered.}
\label{tab:metre-pattern}
\begin{tabular}{llcccc}
\toprule
\textbf{Language} & \textbf{Set} & \textbf{BERT} & \textbf{DistilBERT} & \textbf{RoBERTa} & \textbf{Metaphor RoBERTa} \\
\midrule

\multirow{3}{*}{EN} 
& Poetry + Novel     & 0.98+0.97 & 0.98+0.97 & 0.97+0.97 & 0.97+0.97 \\
& Poetry + Drama     & 0.97+0.90 & 0.97+0.91 & 0.97+0.89 & 0.97+0.90 \\
& Novel + Drama      & 0.91+0.68 & 0.89+0.68 & 0.90+0.58 & 0.87+0.58 \\
\cmidrule(lr){1-6}

\multirow{3}{*}{FR} 
& Poetry + Novel     & 0.71+0.74 & 0.75+0.69 & 0.78+0.83 & 0.76+0.82 \\
& Poetry + Drama     & 0.74+0.70 & 0.77+0.73 & 0.73+0.67 & 0.77+0.73 \\
& Novel + Drama      & 0.74+0.69 & 0.72+0.70 & 0.75+0.73 & 0.74+0.73 \\
\cmidrule(lr){1-6}

\multirow{3}{*}{DE} 
& Poetry + Novel     & 0.73+0.79 & 0.74+0.77 & 0.79+0.83 & 0.80+0.85 \\
& Poetry + Drama     & 0.75+0.70 & 0.74+0.70 & 0.73+0.72 & 0.76+0.73 \\
& Novel + Drama      & 0.78+0.73 & 0.76+0.73 & 0.80+0.76 & 0.70+0.66 \\
\cmidrule(lr){1-6}

\multirow{3}{*}{ES} 
& Poetry + Novel     & 0.79+0.81 & 0.78+0.77 & 0.80+0.82 & 0.80+0.82 \\
& Poetry + Drama     & 0.77+0.70 & 0.75+0.71 & 0.72+0.69 & 0.75+0.74 \\
& Novel + Drama      & 0.76+0.68 & 0.76+0.66 & 0.66+0.54 & 0.80+0.70 \\
\cmidrule(lr){1-6}

\multirow{3}{*}{IT} 
& Poetry + Novel     & 0.80+0.83 & 0.72+0.80 & 0.75+0.83 & 0.81+0.85 \\
& Poetry + Drama     & 0.79+0.75 & 0.79+0.75 & 0.80+0.79 & 0.82+0.78 \\
& Novel + Drama      & 0.79+0.71 & 0.80+0.74 & 0.82+0.67 & 0.81+0.69 \\
\cmidrule(lr){1-6}

\multirow{3}{*}{PT} 
& Poetry + Novel     & 0.81+0.85 & 0.80+0.84 & 0.62+0.78 & 0.86+0.89 \\
& Poetry + Drama     & 0.76+0.60 & 0.73+0.63 & 0.73+0.62 & 0.76+0.70 \\
& Novel + Drama      & 0.78+0.62 & 0.77+0.63 & 0.78+0.68 & 0.70+0.63 \\
\cmidrule(lr){1-6}

\multirow{3}{*}{Average} 
 & Poetry + Novel    & 0.80+0.83 & 0.79+0.81 & 0.79+0.84 & 0.83+0.87 \\
 & Poetry + Drama    & 0.80+0.72 & 0.79+0.74 & 0.78+0.73 & 0.81+0.76 \\
 & Novel + Drama     & 0.79+0.69 & 0.78+0.69 & 0.79+0.66 & 0.77+0.66 \\
\bottomrule

\end{tabular}
\end{table*}

\end{document}